\documentclass[10pt]{article} % For LaTeX2e
\usepackage[preprint]{tmlr}

\usepackage{amsmath,amsfonts,bm}

\def\eqref#1{equation~\ref{#1}}
\def\1{\bm{1}}

\DeclareMathAlphabet{\mathsfit}{\encodingdefault}{\sfdefault}{m}{sl}
\SetMathAlphabet{\mathsfit}{bold}{\encodingdefault}{\sfdefault}{bx}{n}

\usepackage{hyperref}
\usepackage{url}

\usepackage{amsmath,amsthm,amssymb}

\usepackage{mathtools}
\usepackage{subcaption}
\usepackage{enumitem}
\usepackage{algorithm}
\usepackage{algorithmic}
\usepackage{booktabs}
\usepackage{multirow}
\usepackage{bbm}
\usepackage{adjustbox}

\numberwithin{equation}{section}

\theoremstyle{assumption}

\theoremstyle{definition}

\theoremstyle{remark}

\usepackage[most]{tcolorbox}
\usepackage{xcolor}

\definecolor{promptgray}{RGB}{242,242,242}
\definecolor{prompttitle}{RGB}{225,225,225}
\definecolor{promptline}{RGB}{180,180,180}

\newtcolorbox{promptbox}[1][]{
  enhanced,
  breakable,
  colback=promptgray,
  colframe=promptline,
  coltitle=black,
  fonttitle=\bfseries,
  title=#1,
  boxrule=0.8pt,
  arc=1mm,
  left=10pt,
  right=10pt,
  top=8pt,
  bottom=8pt,
  boxsep=0pt,
  toptitle=6pt,
  bottomtitle=6pt,
  colbacktitle=prompttitle,
}

\title{How Prompts Move Language Model Behavior:\\Frames, Salience, and Construal as Semantic Control}

\author{\name Dongseok Kim\thanks{These authors contributed equally.} \email jkds5920@gachon.ac.kr \\
      \addr Department of Computer Engineering\\
      Gachon University
      \AND
      \name Hyoungsun Choi\footnotemark[1] \email hschoi@gachon.ac.kr \\
      \addr Department of Computer Engineering\\
      Gachon University
      \AND
      \name Mohamed Jismy Aashik Rasool\footnotemark[1] \email aashikrasool@gachon.ac.kr \\
      \addr Department of Computer Engineering\\
      Gachon University
      \AND
      \name Gisung Oh\thanks{Co-corresponding authors.} \email eustia@gachon.ac.kr \\
      \addr Department of Computer Engineering\\
      Gachon University}

\def\month{MM}  % Insert correct month for camera-ready version
\def\year{YYYY} % Insert correct year for camera-ready version
\def\openreview{\url{https://openreview.net/forum?id=XXXX}} % Insert correct link to OpenReview for camera-ready version

\begin{document}

\maketitle

\begin{abstract}
Prompt engineering is widely used to shape large language model behavior, yet it is often treated as a practical heuristic rather than as a form of natural-language control. This paper develops a cognitive-semantic account in which prompts function as semantic conditions on how a fixed model interprets inputs, foregrounds information, and structures tasks. We formalize this account through three notions---frame activation, salience control, and construal selection---and study them in natural language inference, claim verification, and multi-hop question answering. Across these settings, prompts produce measurable changes in label judgments, evidence use, and answer--support organization, showing that prompt effects differ not only in magnitude but also in semantic direction. The paper therefore reframes prompting as the analysis of how instructions move model behavior, rather than only whether they improve performance.
\end{abstract}

\section{Introduction}
\label{sec:introduction}

Prompt engineering has become one of the main ways of shaping large language model behavior. In many settings, users do not alter a model's parameters, training data, or decoding algorithm; they change the natural-language instructions through which a task is presented. The same input can thereby be treated as a classification problem, a verification problem, a reasoning problem, a comparison problem, or an evidence-selection problem. Prompting therefore does more than request an output: it helps specify the semantic conditions under which an input is interpreted and made actionable for the model.

Existing work has shown that language models are sensitive to prompt wording, formatting, demonstrations, and task presentation. Yet prompt sensitivity alone does not explain what kind of behavioral change has occurred. Two prompts may produce different outputs, but the relevant question is whether the prompt moves the model toward a particular judgment, foregrounds a particular kind of evidence, or reorganizes the task around a different response structure. A theory of prompting should therefore describe not only the magnitude of prompt-induced variation, but also its semantic direction.

This paper develops a cognitive-semantic account of prompting as semantic control. The central claim is that prompts shape model behavior by organizing how an input is interpreted, what information becomes salient, and how the task is construed. A prompt can activate an interpretive frame, foreground selected parts of the input, or make the same task appear as a direct-answer problem, an evidence-first problem, a bridge-reasoning problem, or a comparison problem. In this view, prompting is a linguistic mechanism for controlling the semantic organization of model behavior.

This perspective shifts the analysis of prompting from prompt quality to prompt effect. A prompt should not be evaluated only by whether it improves aggregate performance. It may instead induce a systematic behavioral movement toward a label, a type of evidence, a supporting structure, or a task interpretation. Such movement can be meaningful even without a performance gain: a prompt that increases contradiction sensitivity, foregrounds refuting evidence, or changes answer--support coupling is not merely better or worse than another prompt, but changes the form of behavior that the model exhibits.

We organize this account around three concepts from cognitive semantics. \emph{Frame activation} captures how prompts place an input under a particular interpretive structure, making certain relations or judgments more prominent. \emph{Salience control} captures how prompts foreground selected information and make it functionally central to the response. \emph{Construal selection} captures how prompts make the same input appear as a different kind of task, thereby changing the organization of the output. Together, these concepts provide a vocabulary for distinguishing prompt effects that are otherwise grouped together as prompt sensitivity.

We operationalize this framework through three experiments. In natural language inference, frame-oriented prompts test whether semantic relation judgments shift across entailment, contradiction, and neutral labels. In claim verification, salience-oriented prompts test whether evidence selection and label--evidence coupling change. In multi-hop question answering, construal-oriented prompts test whether different task framings alter the relationship between final answers and supporting facts. Across all three settings, the input examples remain fixed while only the prompt condition varies, allowing us to measure how natural-language instructions move model behavior across judgments, evidential priorities, and task structures.

Our contributions are as follows:
\begin{itemize}
    \item \textbf{A cognitive-semantic account of prompting.}
    Prompt engineering is reinterpreted as semantic control over a fixed language model, where prompts shape how an input is interpreted, foregrounded, and organized.

    \item \textbf{A shift from prompt quality to prompt effect.}
    The analysis focuses on the direction and structure of prompt-induced behavioral movement, rather than treating performance improvement as the only relevant outcome.

    \item \textbf{An operational framework for semantic prompt effects.}
    Frame activation, salience control, and construal selection are introduced as three measurable forms of prompt-induced behavioral variation.

    \item \textbf{A three-part empirical evaluation.}
    Experiments on natural language inference, claim verification, and multi-hop question answering show how prompts move model behavior across labels, evidence selection, and answer--support relations.
\end{itemize}

\section{Related Work}
\label{sec:related_work}

\subsection{Prompting and Instructional Control}
\label{subsec:prompting}

Prompting has become a central interface for controlling language model behavior without modifying model parameters at inference time. Early work showed that pretrained language models can be adapted to new tasks by changing the textual form of the input, using natural-language prompts, cloze-style formulations, or automatically generated trigger tokens~\citep{brown2020language, shin2020autoprompt, liu2023pre}. This control was later made more systematic through discrete prompt optimization, continuous prompt representations, and instruction tuning over broad collections of natural-language tasks~\citep{gao2021making, li2021prefix, lester2021power, wei2021finetuned, sanh2021multitask, mishra2022cross, wang2022super, chung2024scaling}.

Prompting also affects the form of model outputs. Chain-of-thought and zero-shot reasoning prompts can elicit intermediate reasoning traces, while instruction-following methods trained with human feedback align outputs more closely with user intent~\citep{wei2022chain, kojima2022large, wang2022self, ouyang2022training}. These approaches establish prompting as a practical control mechanism, but they primarily study how prompts improve task performance, elicit reasoning, or make models follow instructions. The present work focuses instead on the semantic dimensions along which prompts move model behavior.

\subsection{Prompt Sensitivity and Behavioral Variation}
\label{subsec:sensitivity}

Prompted language model behavior is highly sensitive to task presentation. In few-shot and in-context learning, predictions vary with prompt format, answer verbalization, label priors, calibration choices, and the selection, order, or composition of demonstrations~\citep{zhao2021calibrate, holtzman2021surface, webson2022prompt, han2022prototypical, lu2022fantastically, min2022rethinking, rubin2022learning, zhang2022active, wu2023self, ye2023compositional, wang2023large}. These findings show that prompts do not merely specify a task; they influence the task representation inferred by the model.

Prompt-induced variation also appears under changes that preserve the nominal task. Models can be distracted by irrelevant context, fail to use information robustly when its position changes, and show large performance differences under formatting changes or adversarial prompt perturbations~\citep{shi2023large, liu2024lost, sclar2023quantifying, zhu2023promptrobust}. This literature motivates a behavioral view of prompting, where the object of analysis is not only model accuracy but also the structure of the variation produced by different prompt conditions.

\subsection{Cognitive Semantics as a Framework for Prompt Effects}
\label{subsec:cognitive_semantics}

Cognitive semantics provides a vocabulary for describing how linguistic form shapes interpretation. Frame semantics treats expressions as being understood relative to structured background situations, while cognitive grammar treats meaning as conceptualization, where the same content can be profiled, focused, and construed in different ways~\citep{petruck1996frame, langacker1986introduction}. Work on construal and salience further shows how alternative formulations can affect which aspects of an event are foregrounded and how the event is organized in interpretation~\citep{divjak2020construal, schmid2016toward}.

Computational semantics has operationalized related ideas through FrameNet, semantic role labeling, and frame-semantic parsing~\citep{baker1998berkeley, gildea2002automatic, das2010probabilistic, das2014frame}. Subsequent work expanded these representations through broader context, heterogeneous annotations, structured frame relations, graded semantic roles, definitional information, and evaluations of frame-semantic competence in modern language models~\citep{roth2015context, kshirsagar2015frame, yang2017joint, reisinger2015semantic, jiang2021exploiting, chundru2025llms}. These traditions make it possible to describe prompt effects in terms of frame activation, salience control, and construal selection, rather than treating prompt sensitivity as an undifferentiated source of behavioral instability.

\subsection{Evidence Grounding and Salience in Model Outputs}
\label{subsec:evidence_salience}

Evidence grounding is central to tasks in which model outputs must be supported by identifiable parts of the input or an external corpus. Fact verification benchmarks require systems to classify claims while retrieving or selecting supporting evidence, and knowledge-intensive benchmarks evaluate whether outputs are grounded in retrieved information and identifiable provenance~\citep{thorne2018fever, petroni2021kilt}. Related work on rationales and salient input regions studies whether selected evidence aligns with predictive or human-interpretable justifications~\citep{lei2016rationalizing, chen2018learning, bastings2019interpretable, deyoung2020eraser}.

At the same time, attention and saliency methods show that apparent importance signals do not always provide faithful explanations of model decisions~\citep{jain2019attention, serrano2019attention, atanasova2020diagnostic}. Retrieval-augmented and citation-supported generation extends this concern to large language models by evaluating whether generated claims are supported by identifiable sources~\citep{guu2020retrieval, lewis2020retrieval, karpukhin2020dense, rashkin2023measuring, gao2023enabling, min2023factscore}. This grounding literature separates the correctness of an output from the evidential structure that supports it, which is essential for analyzing how prompts alter what information becomes central to a model's response.

\subsection{Multi-Hop Reasoning and Task Construal}
\label{subsec:task_construal}

Multi-hop question answering studies whether models can combine information distributed across multiple facts, documents, or modalities. Early datasets framed this problem as reading comprehension across documents, while later analyses showed that some benchmarks contain shortcuts or design choices that allow models to answer without performing the intended reasoning process~\citep{welbl2018constructing, talmor2018web, yang2018hotpotqa, chen2019understanding, min2019compositional, jiang2019avoiding}.

A second line of work makes the intermediate organization of reasoning more explicit through question decomposition, reasoning-path rescoring, structured decompositions, and graphs over entities, sentences, paragraphs, or documents~\citep{min2019multi, wolfson2020break, ding2019cognitive, qiu2019dynamically, nishida2019answering, fang2020hierarchical}. Recent benchmarks broaden the forms of organization required for complex question answering, including table-and-text reasoning, implicit strategy questions, and compositionally constructed multi-hop questions~\citep{chen2020open, geva2021did, trivedi2022musique}. This work motivates treating task framing as a source of behavioral variation: the same question-answering input can be organized as a direct-answer task, an evidence-first task, a bridge-reasoning task, or a comparison task.

\section{Theory}
\label{sec:theory}

\subsection{Prompting as Semantic Control}
\label{subsec:prompting_semantic_control}

We treat prompting as semantic control over a fixed language model. In the setting considered here, the model parameters, task input, decoding procedure, and output format are held fixed, while only the natural-language prompt varies. This setup isolates the prompt as the source of behavioral change: when the underlying input remains the same, the prompt changes how that input is interpreted, what information is prioritized, and what kind of response is made appropriate.

A prompt is therefore more than an additional string attached to a task instance. It specifies the relation between the input, the task, and the expected response. The same premise--hypothesis pair can be judged under an entailment-oriented or contradiction-oriented frame; the same claim can be verified by foregrounding different evidence; the same question-answering context can be organized as direct answering, bridge reasoning, support selection, or comparison. In each case, the prompt changes the semantic organization of the task without changing the task input itself.

This control is visible as systematic behavioral movement. Prompts can shift label distributions, change class-wise recall, alter evidence selection, affect supporting-fact choice, or modify the coupling between an answer and its support. We analyze these effects through three concepts from cognitive semantics: frame activation, salience control, and construal selection. These concepts mark different levels at which prompts organize model behavior: the criteria of judgment, the foregrounding of information, and the structure of the task.

\subsection{Frame Activation in Semantic Judgment}
\label{subsec:frames}

Frame activation concerns the interpretive structure under which an input is judged. In cognitive semantics, a frame makes certain aspects of a situation relevant for interpretation: it specifies what kind of situation is being considered, which relations matter, and what counts as an appropriate judgment. Applied to prompting, frame activation means that a prompt can make one semantic relation more prominent than another while leaving the input unchanged.

Natural language inference provides a direct setting for this form of control. A premise--hypothesis pair may be presented under a neutral relation-judgment frame, an entailment-oriented frame, a contradiction-oriented frame, or an uncertainty-oriented frame. The entailment-oriented prompt foregrounds support, the contradiction-oriented prompt foregrounds incompatibility, and the uncertainty-oriented prompt foregrounds missing information or underdetermination. The model is still asked to judge the same pair, but the prompt changes which relation is made most available as the basis for judgment.

The expected effect is directional rather than merely variable. An entailment frame should move behavior toward entailment-sensitive decisions, a contradiction frame toward contradiction-sensitive decisions, and an uncertainty frame toward neutral or underdetermined judgments. We measure this movement through changes in NLI predictions, frame-compatible class recall, and shifts in the model's label distribution toward the relation foregrounded by the prompt.

\subsection{Salience Control in Evidence Selection}
\label{subsec:salience}

Salience control concerns how information is foregrounded within a task. Interpretation depends not only on what information is available, but also on which parts of that information become functionally central to the response. This distinction matters when an input contains multiple candidate evidence sentences, only some of which should guide the final decision.

Claim verification makes this form of control measurable. The model receives the same claim and the same evidence candidates, while the prompt changes the evidential priority of the task. A neutral verification prompt asks for a label decision. An evidence-first prompt makes evidence identification prior to labeling. A support-oriented prompt foregrounds confirming evidence, while a refutation-oriented prompt foregrounds contradicting evidence. These prompts can therefore change both the selected evidence and the label supported by that evidence.

The relevant movement is not exhausted by label accuracy. A prompt may improve or degrade the final label while also changing whether the model selects gold evidence, whether label prediction and evidence selection succeed together, and whether distractor evidence enters the rationale. We therefore analyze salience control through evidence F1, label--evidence joint performance, and distractor leakage, alongside ordinary verification accuracy.

\subsection{Construal Selection in Task Interpretation}
\label{subsec:construal}

Construal selection concerns the organization of the task as a whole. Whereas frames guide the criteria for judgment and salience guides the foregrounding of information, construal determines what kind of problem the model treats the input as posing. The same content can be construed as a direct question, a comparison, a sequence of reasoning steps, or a problem requiring an intermediate link.

Multi-hop question answering provides a setting in which this task-level organization becomes visible. The same question and context can be presented as a direct answering task, a supporting-facts-first task, a bridge-reasoning task, or a comparison task. A direct-answer prompt centers the final response. A supporting-facts-first prompt organizes the task around the evidential path. A bridge prompt makes intermediate entities or linking facts central, while a comparison prompt foregrounds distinctions or relations between entities.

The main object of analysis is the relation between answers and supporting structure. A model can produce the correct answer without selecting the appropriate supporting facts, or identify relevant facts without producing the correct final answer. Construal selection therefore cannot be captured by answer accuracy alone. We measure it through answer quality, supporting-fact overlap, joint answer--support performance, and type-matched construal effects in multi-hop question answering.

\section{Experiments}
\label{sec:experiments}

\subsection{Common Experimental Design}
\label{subsec:common_experimental_design}

The experiments measure prompt-induced behavioral movement under controlled prompt variation. In each setting, the benchmark example, model, decoding procedure, and output schema are held fixed, while only the natural-language prompt changes. This design allows differences in label choice, evidence selection, supporting-fact choice, and answer--support coupling to be analyzed as changes in how the same input is semantically organized by the model.

The three benchmarks correspond to the three levels of semantic control introduced above. SNLI tests frame activation through semantic relation judgments over premise--hypothesis pairs. FEVER tests salience control by separating claim labels from the evidence selected to support them. HotpotQA tests construal selection by requiring both final answers and supporting facts in multi-hop question answering. We evaluate SmolLM2-1.7B-Instruct, Qwen3-4B-Instruct, and Mistral-7B-Instruct-v0.3. All responses are constrained to task-specific JSON formats so that labels, evidence identifiers, answers, and supporting sentence identifiers can be parsed automatically. Format validity and extraction rates are recorded as implementation diagnostics; the main analysis focuses on parsed task behavior.

The evaluation combines standard task metrics with prompt-effect diagnostics. Standard metrics include label accuracy, answer exact match, answer F1, evidence F1, and supporting-fact F1. These scores measure ordinary benchmark performance. Prompt-effect diagnostics compare semantic prompt conditions with neutral or direct baselines, measuring whether a prompt moves behavior toward a frame-compatible label, a different evidential basis, or a different answer--support structure. Full metric definitions are provided in Appendix~\ref{app:metric_definitions}; the main diagnostics are defined below.

For the frame experiment, frame sensitivity measures whether an example receives different labels under different prompt frames:
\[
\mathrm{FrameSens}
=
\frac{1}{N}
\sum_{i=1}^{N}
\mathbf{1}
\left[
\exists a,b:\hat{y}_{i}^{(a)} \neq \hat{y}_{i}^{(b)}
\right].
\]
Target-frame recall gain measures whether each frame increases recall for its corresponding label relative to the neutral prompt:
\[
\mathrm{FrameRecGain}
=
\frac{1}{|C|}
\sum_{c \in C}
\left(
\mathrm{Recall}_{c}(p_c)
-
\mathrm{Recall}_{c}(p_{\mathrm{neutral}})
\right).
\]
Frame-specific bias shift measures the corresponding change in prediction rate:
\[
\mathrm{BiasShift}_{c}
=
P(\hat{y}=c \mid p_c)
-
P(\hat{y}=c \mid p_{\mathrm{neutral}}).
\]
Together, these diagnostics distinguish undirected prompt sensitivity from movement toward the relation foregrounded by the prompt.

For the FEVER experiment, salience gain measures whether an evidence-first prompt increases evidence overlap relative to the neutral verification prompt:
\[
\mathrm{SalGain}
=
\mathrm{EvidF1}(p_{\mathrm{evidence}})
-
\mathrm{EvidF1}(p_{\mathrm{neutral}}).
\]
Joint gain measures the corresponding change in label--evidence joint performance:
\[
\mathrm{JointGain}
=
\mathrm{Joint}(p_{\mathrm{evidence}})
-
\mathrm{Joint}(p_{\mathrm{neutral}}).
\]
Leakage reduction measures whether the evidence-first prompt reduces the selection of distractor evidence:
\[
\mathrm{LeakReduction}
=
\mathrm{Leak}(p_{\mathrm{neutral}})
-
\mathrm{Leak}(p_{\mathrm{evidence}}).
\]
These quantities track how prompts change the evidential basis of verification, rather than only the final label.

For the HotpotQA experiment, joint answer--support performance is defined at the example level as
\[
\mathrm{JointF1}_{i}
=
\mathrm{AnsF1}_{i}
\cdot
\mathrm{SupF1}_{i}.
\]
This score is high only when the model both answers the question and selects appropriate supporting facts. Type-matched construal gain compares bridge prompts on bridge questions and comparison prompts on comparison questions against the direct-answer baseline:
\[
\begin{aligned}
\mathrm{TypeMatchGain}
&=
\frac{1}{|\mathcal{B}|}
\sum_{i \in \mathcal{B}}
\left(
\mathrm{JointF1}_{i}(p_{\mathrm{bridge}})
-
\mathrm{JointF1}_{i}(p_{\mathrm{direct}})
\right) \\
&\quad +
\frac{1}{|\mathcal{M}|}
\sum_{i \in \mathcal{M}}
\left(
\mathrm{JointF1}_{i}(p_{\mathrm{comparison}})
-
\mathrm{JointF1}_{i}(p_{\mathrm{direct}})
\right),
\end{aligned}
\]
where \(\mathcal{B}\) denotes bridge questions and \(\mathcal{M}\) denotes comparison questions. Bootstrap confidence intervals for the main aggregate metrics are reported in Appendix~\ref{app:experiment_confidence_intervals}.

\subsection{Experiment 1: Prompt Frames in Natural Language Inference}
\label{subsec:exp1}

\paragraph{Experimental setup.}
The first experiment examines whether prompt frames shift semantic relation judgments in natural language inference. Each SNLI example is presented with the same premise--hypothesis pair across four prompt conditions: a neutral NLI prompt, an entailment-oriented prompt, a contradiction-oriented prompt, and an uncertainty-oriented prompt. The model returns one of three labels: entailment, contradiction, or neutral. Because the input pair and output space remain fixed, differences across conditions can be attributed to the interpretive frame introduced by the prompt.

\paragraph{Results and analysis.}
Table~\ref{tab:exp1_snli_results} shows clear frame effects, with different models responding to the prompt frames in different ways. SmolLM2 exhibits the strongest movement: its predictions often shift toward the label foregrounded by the prompt, but this responsiveness reduces balanced classification performance in several conditions. Qwen remains comparatively stable while still showing targeted shifts, especially under contradiction and uncertainty frames. Mistral occupies an intermediate position, with visible movement toward contradiction- and uncertainty-compatible judgments. The frame-level diagnostics indicate that these changes are not merely undirected prompt sensitivity; the prompts tend to move predictions toward the semantic relation they foreground.

\begin{table*}[ht]
\centering
\small
\setlength{\tabcolsep}{4pt}
\caption{Experiment 1 point estimates on SNLI.}
\label{tab:exp1_snli_results}

\begin{subtable}{\textwidth}
\centering
\caption{Prompt-condition results.}
\begin{tabular}{llrrrrr}
\toprule
\multicolumn{1}{c}{\multirow{2}{*}{Model}} &
\multicolumn{1}{c}{\multirow{2}{*}{Condition}} &
\multicolumn{1}{c}{\multirow{2}{*}{Accuracy}} &
\multicolumn{1}{c}{\multirow{2}{*}{Macro-F1}} &
\multicolumn{3}{c}{Recall} \\
\cmidrule(lr){5-7}
& & & &
\multicolumn{1}{c}{Entailment} &
\multicolumn{1}{c}{Contradiction} &
\multicolumn{1}{c}{Neutral} \\
\midrule
\multirow{4}{*}{SmolLM2-1.7B}
& contradiction & 0.458 & 0.365 & 0.922 & 0.446 & 0.006 \\
& entailment    & 0.378 & 0.259 & 0.994 & 0.078 & 0.060 \\
& neutral       & 0.476 & 0.437 & 0.916 & 0.289 & 0.222 \\
& uncertainty   & 0.422 & 0.367 & 0.665 & 0.072 & 0.527 \\
\midrule
\multirow{4}{*}{Qwen3-4B}
& contradiction & 0.864 & 0.864 & 0.778 & 0.970 & 0.844 \\
& entailment    & 0.830 & 0.832 & 0.880 & 0.693 & 0.916 \\
& neutral       & 0.858 & 0.860 & 0.874 & 0.807 & 0.892 \\
& uncertainty   & 0.754 & 0.756 & 0.754 & 0.536 & 0.970 \\
\midrule
\multirow{4}{*}{Mistral-7B}
& contradiction & 0.710 & 0.703 & 0.898 & 0.452 & 0.778 \\
& entailment    & 0.614 & 0.534 & 0.892 & 0.054 & 0.892 \\
& neutral       & 0.624 & 0.581 & 0.910 & 0.187 & 0.772 \\
& uncertainty   & 0.520 & 0.438 & 0.581 & 0.012 & 0.964 \\
\bottomrule
\end{tabular}
\end{subtable}

\vspace{0.75em}

\begin{subtable}{\textwidth}
\centering
\caption{Frame-level results.}
\resizebox{\textwidth}{!}{
\begin{tabular}{lrrrrrrrr}
\toprule
\multicolumn{1}{c}{\multirow{2}{*}{Model}} &
\multicolumn{1}{c}{\multirow{2}{*}{FrameSens}} &
\multicolumn{1}{c}{\multirow{2}{*}{FrameRecGain}} &
\multicolumn{3}{c}{Gain} &
\multicolumn{3}{c}{Shift} \\
\cmidrule(lr){4-6}
\cmidrule(lr){7-9}
& & &
\multicolumn{1}{c}{Entailment} &
\multicolumn{1}{c}{Contradiction} &
\multicolumn{1}{c}{Neutral} &
\multicolumn{1}{c}{Entailment} &
\multicolumn{1}{c}{Contradiction} &
\multicolumn{1}{c}{Neutral} \\
\midrule
SmolLM2-1.7B & 0.570 & 0.180 & 0.078 & 0.157 & 0.305 & 0.238 & 0.134 & 0.266 \\
Qwen3-4B    & 0.244 & 0.082 & 0.006 & 0.163 & 0.078 & 0.002 & 0.080 & 0.162 \\
Mistral-7B  & 0.350 & 0.146 & -0.018 & 0.265 & 0.192 & -0.046 & 0.104 & 0.236 \\
\bottomrule
\end{tabular}
}
\end{subtable}

\end{table*}

\subsection{Experiment 2: Evidence Foregrounding in Claim Verification}
\label{subsec:exp2}

\paragraph{Experimental setup.}
The second experiment studies salience control in FEVER claim verification. Each example contains a claim and a fixed set of candidate evidence sentences, including gold evidence when available and distractor sentences sampled from the corpus. The prompt varies the evidential priority of the task: neutral verification, evidence-first selection, support-oriented foregrounding, or refutation-oriented foregrounding. The model returns both a verification label and selected evidence sentence identifiers, which makes it possible to evaluate label prediction, evidence selection, and their coupling within the same output.

\paragraph{Results and analysis.}
Tables~\ref{tab:exp2_fever_condition_results} and~\ref{tab:exp2_fever_salience_results} show that foregrounding instructions affect both the selected evidence and the associated label behavior. SmolLM2 changes its evidence-selection pattern and label distribution across prompts, although its evidence choices remain noisy. Qwen achieves the strongest neutral verification performance, and the salience prompts mainly produce targeted changes in evidence overlap, leakage, and class-wise recall rather than broad performance gains. Mistral displays the most interpretable salience pattern: evidence-first and support-oriented prompts improve several grounding-related measures, while refutation-oriented prompting shifts behavior toward REFUTES decisions. These results support the view that salience prompts do not simply make verification better or worse; they change which information becomes central to the model's decision.

\begin{table*}[ht]
\centering
\small
\setlength{\tabcolsep}{3pt}
\caption{Condition-level point estimates for Experiment 2.}
\label{tab:exp2_fever_condition_results}
\resizebox{\textwidth}{!}{
\begin{tabular}{llrrrrrrrr}
\toprule
\multicolumn{1}{c}{\multirow{2}{*}{Model}} &
\multicolumn{1}{c}{\multirow{2}{*}{Condition}} &
\multicolumn{1}{c}{\multirow{2}{*}{LabelAcc}} &
\multicolumn{1}{c}{\multirow{2}{*}{Macro-F1}} &
\multicolumn{1}{c}{\multirow{2}{*}{EvidF1}} &
\multicolumn{1}{c}{\multirow{2}{*}{Joint}} &
\multicolumn{1}{c}{\multirow{2}{*}{Leak}} &
\multicolumn{3}{c}{Recall} \\
\cmidrule(lr){8-10}
& & & & & & &
\multicolumn{1}{c}{SUPPORTS} &
\multicolumn{1}{c}{REFUTES} &
\multicolumn{1}{c}{NEI} \\
\midrule
\multirow{4}{*}{SmolLM2-1.7B}
& neutral        & 0.440 & 0.336 & 0.102 & 0.036 & 0.909 & 0.940 & 0.000 & 0.380 \\
& evidence-first & 0.530 & 0.429 & 0.128 & 0.022 & 0.913 & 0.641 & 0.000 & 0.952 \\
& support        & 0.386 & 0.267 & 0.091 & 0.038 & 0.909 & 0.976 & 0.000 & 0.181 \\
& refutation     & 0.518 & 0.415 & 0.122 & 0.026 & 0.907 & 0.719 & 0.000 & 0.837 \\
\midrule
\multirow{4}{*}{Qwen3-4B}
& neutral        & 0.920 & 0.920 & 0.907 & 0.910 & 0.022 & 0.922 & 0.916 & 0.922 \\
& evidence-first & 0.902 & 0.902 & 0.883 & 0.894 & 0.010 & 0.934 & 0.802 & 0.970 \\
& support        & 0.896 & 0.895 & 0.888 & 0.888 & 0.016 & 0.940 & 0.808 & 0.940 \\
& refutation     & 0.910 & 0.911 & 0.913 & 0.900 & 0.032 & 0.916 & 0.940 & 0.873 \\
\midrule
\multirow{4}{*}{Mistral-7B}
& neutral        & 0.598 & 0.539 & 0.601 & 0.592 & 0.010 & 0.707 & 0.108 & 0.982 \\
& evidence-first & 0.610 & 0.547 & 0.620 & 0.608 & 0.008 & 0.749 & 0.102 & 0.982 \\
& support        & 0.614 & 0.536 & 0.618 & 0.610 & 0.014 & 0.814 & 0.060 & 0.970 \\
& refutation     & 0.544 & 0.519 & 0.522 & 0.504 & 0.084 & 0.377 & 0.299 & 0.958 \\
\bottomrule
\end{tabular}
}
\end{table*}

\begin{table*}[ht]
\centering
\small
\setlength{\tabcolsep}{3pt}
\caption{Salience-level point estimates for Experiment 2. Verification denotes metrics computed only on verifiable examples with SUPPORTS or REFUTES labels.}
\label{tab:exp2_fever_salience_results}
\resizebox{\textwidth}{!}{
\begin{tabular}{lrrrrrrrrrr}
\toprule
\multicolumn{1}{c}{\multirow{2}{*}{Model}} &
\multicolumn{1}{c}{\multirow{2}{*}{SalGain}} &
\multicolumn{1}{c}{\multirow{2}{*}{JointGain}} &
\multicolumn{1}{c}{\multirow{2}{*}{LeakRed.}} &
\multicolumn{2}{c}{Recall Gain} &
\multicolumn{2}{c}{Bias Shift} &
\multicolumn{3}{c}{Verification} \\
\cmidrule(lr){5-6}
\cmidrule(lr){7-8}
\cmidrule(lr){9-11}
& & & &
\multicolumn{1}{c}{SUPPORTS} &
\multicolumn{1}{c}{REFUTES} &
\multicolumn{1}{c}{SUPPORTS} &
\multicolumn{1}{c}{REFUTES} &
\multicolumn{1}{c}{SalGain} &
\multicolumn{1}{c}{JointGain} &
\multicolumn{1}{c}{LeakRed.} \\
\midrule
SmolLM2-1.7B & 0.026 & -0.014 & -0.004 & 0.036 & 0.000 & 0.144 & 0.000 & 0.039 & -0.021 & -0.006 \\
Qwen3-4B    & -0.024 & -0.016 & 0.012 & 0.018 & 0.024 & 0.020 & 0.034 & -0.051 & -0.048 & 0.003 \\
Mistral-7B  & 0.020 & 0.016 & 0.002 & 0.108 & 0.192 & 0.040 & 0.134 & 0.029 & 0.024 & 0.003 \\
\bottomrule
\end{tabular}
}
\end{table*}

\subsection{Experiment 3: Task Construal in Multi-Hop Question Answering}
\label{subsec:exp3}

\paragraph{Experimental setup.}
The third experiment evaluates task construal in HotpotQA. We use a balanced sample of bridge and comparison questions, each requiring a final answer and supporting facts. For every example, the question and candidate context sentences are held fixed while the prompt presents the task as direct answering, supporting-facts-first answering, bridge reasoning, or comparison reasoning. The model returns a final answer together with selected supporting sentence identifiers. This design tests whether changing the construal of the task alters answer generation, supporting-fact selection, or the relation between the two.

\paragraph{Results and analysis.}
Tables~\ref{tab:exp3_hotpot_condition_results} and~\ref{tab:exp3_hotpot_construal_results} show that construal prompts have weaker and less uniform effects than the frame and salience prompts. SmolLM2 remains low in answer quality across conditions, with most observable variation appearing in supporting-fact behavior. Qwen performs best overall, but support-first and bridge prompts mainly adjust support selection and answer--support coupling relative to the direct-answer baseline. Mistral follows a similar pattern: final-answer quality changes little, while support-related behavior is more responsive to the prompt. The main effect of construal prompting is therefore organizational rather than purely accuracy-oriented. It changes how answers and supporting facts are coordinated, but the direction of that change is more model- and task-type-dependent than in the previous experiments.

\begin{table*}[ht]
\centering
\small
\setlength{\tabcolsep}{3pt}
\caption{Condition-level point estimates for Experiment 3.}
\label{tab:exp3_hotpot_condition_results}
\begin{tabular}{llrrrrrr}
\toprule
Model & Condition & AnsEM & AnsF1 & SupF1 & JointF1 & Coupling & PredSup \\
\midrule
\multirow{4}{*}{SmolLM2-1.7B}
& direct       & 0.050 & 0.050 & 0.192 & 0.011 & 0.300 & 3.230 \\
& support-first& 0.050 & 0.050 & 0.185 & 0.010 & 0.200 & 3.185 \\
& bridge       & 0.050 & 0.050 & 0.196 & 0.010 & 0.200 & 3.265 \\
& comparison   & 0.050 & 0.050 & 0.180 & 0.008 & 0.200 & 2.990 \\
\midrule
\multirow{4}{*}{Qwen3-4B}
& direct       & 0.600 & 0.724 & 0.773 & 0.575 & 0.929 & 1.785 \\
& support-first& 0.595 & 0.716 & 0.792 & 0.575 & 0.936 & 1.810 \\
& bridge       & 0.585 & 0.723 & 0.788 & 0.580 & 0.929 & 1.840 \\
& comparison   & 0.525 & 0.666 & 0.777 & 0.536 & 0.919 & 1.770 \\
\midrule
\multirow{4}{*}{Mistral-7B}
& direct       & 0.425 & 0.541 & 0.636 & 0.380 & 0.926 & 1.960 \\
& support-first& 0.400 & 0.528 & 0.625 & 0.364 & 0.911 & 1.975 \\
& bridge       & 0.430 & 0.541 & 0.619 & 0.375 & 0.917 & 1.965 \\
& comparison   & 0.420 & 0.541 & 0.634 & 0.363 & 0.849 & 1.950 \\
\bottomrule
\end{tabular}
\end{table*}

\begin{table*}[ht]
\centering
\small
\setlength{\tabcolsep}{3pt}
\caption{Construal-level point estimates for Experiment 3. Gains are computed relative to the direct-answer construal.}
\label{tab:exp3_hotpot_construal_results}
\resizebox{\textwidth}{!}{
\begin{tabular}{lrrrrrrrr}
\toprule
Model & SupFirst & TypeMatch & AnsGain & SupGain & SuppExp. & BridgeOnly & CompOnly & CoupleGain \\
\midrule
SmolLM2-1.7B & -0.002 & -0.003 & 0.000 & -0.007 & -0.260 & 0.000 & -0.007 & -0.100 \\
Qwen3-4B    & 0.000 & -0.014 & -0.028 & 0.008 & 0.045 & 0.019 & -0.046 & -0.006 \\
Mistral-7B  & -0.016 & -0.002 & 0.024 & -0.017 & 0.000 & -0.004 & -0.001 & -0.066 \\
\bottomrule
\end{tabular}
}
\end{table*}

\section{Discussion}
\label{sec:discussion}

\subsection{Prompting as Semantic Control}
\label{subsec:semantic_control}

The experiments support the view that prompts act as semantic control conditions rather than as neutral task wrappers. Across natural language inference, claim verification, and multi-hop question answering, the input instances are held fixed while the prompt conditions vary. The resulting changes appear in relation judgments, evidence selection, supporting-fact behavior, and answer--support coupling. A frame-oriented prompt changes the criteria under which a relation is evaluated; a salience-oriented prompt changes which parts of the input become central to a decision; and a construal-oriented prompt changes how the task is organized as a problem to be solved.

This view makes prompt-induced variation an object of analysis in its own right. A prompt does not only make a model more or less accurate; it can make the model more contradiction-sensitive, more evidence-oriented, more cautious, or more structurally explicit. Such effects are visible in label distributions, class-wise recall, evidence overlap, distractor leakage, support selection, and answer--support alignment. These quantities describe how behavior moves under a prompt, not only whether the resulting output receives a higher benchmark score.

\subsection{What the Three Experiments Show}
\label{subsec:experimental_synthesis}

The three experiments reveal different forms of semantic movement. In the NLI experiment, frame-oriented prompts produce the most direct effect: predictions shift toward the semantic relation foregrounded by the prompt. In the FEVER experiment, salience-oriented prompts affect the evidential basis of verification, changing selected evidence, leakage, and label--evidence coupling. In the HotpotQA experiment, construal-oriented prompts produce weaker and more model-dependent changes, mainly in how answers and supporting facts are coordinated.

These differences are important because they show that prompt sensitivity is not a single phenomenon. Frame activation operates on judgment criteria, salience control operates on evidential priority, and construal selection operates on task organization. The experiments therefore separate three ways in which natural-language instructions shape model behavior: changing what relation is judged, changing what information is foregrounded, and changing how a multi-step task is structured.

\subsection{Prompt Effects Beyond Performance Improvement}
\label{subsec:beyond_performance}

The results also show why prompt effects should not be reduced to aggregate performance. Several prompts induce systematic behavioral changes without uniformly improving accuracy, F1, or joint performance. A frame prompt may move predictions toward a frame-compatible label even if macro-level performance decreases. A salience prompt may change the evidence selected for a claim without improving the final verification label. A construal prompt may alter supporting-fact behavior while leaving answer quality nearly unchanged.

This distinction matters for both analysis and system design. If prompts are evaluated only by benchmark scores, then shifts in judgment criteria, evidential basis, or task organization remain hidden. By measuring behavioral direction separately from performance, the framework makes it possible to describe prompts as interventions on the form of model behavior. The relevant question becomes not only which prompt works best, but what kind of behavior each prompt elicits.

\subsection{A Hierarchy of Semantic Control}
\label{subsec:semantic_control_hierarchy}

The empirical pattern suggests a hierarchy of semantic control. Frame activation is the most direct form because it changes the evaluative criterion applied to a fixed input. Salience control is more demanding because the model must identify relevant information among competing candidates and connect it to the final decision. Construal selection is deeper still: it asks the model to reorganize the task as a structured procedure involving intermediate support, reasoning type, or answer--evidence coordination.

This hierarchy explains why the experiments differ in strength and stability. Frame prompts often produce clear directional shifts because the target label relation is explicit. Salience prompts depend on both wording and the evidence environment, making their effects more sensitive to model behavior and context structure. Construal prompts require the model to sustain a task-level organization across multiple output components, so their effects are less uniform and more dependent on the model's ability to coordinate answer generation with support selection.

\subsection{From Prompt Engineering to Context and Harness Engineering}
\label{subsec:prompt_context_harness}

The hierarchy also clarifies the relation between prompt engineering, context engineering, and harness engineering. Frame activation is closest to ordinary prompt engineering: a natural-language instruction can specify a role, criterion, perspective, or label relation. Salience control points toward context engineering, because foregrounding depends on whether relevant information is available, distinguishable from distractors, and placed in a useful context. Construal selection points toward harness engineering, since task-level organization may require decomposition, routing, verification, structured outputs, or tool-mediated intermediate steps.

Prompting is therefore one layer in a broader stack of semantic control. Prompts provide the linguistic entry point, contexts shape the information environment, and harnesses structure the procedure through which complex tasks are executed and checked. The experiments illustrate why these layers cannot be collapsed into prompt wording alone: relation judgments can often be shifted directly through instructions, evidence use depends on the organization of the input context, and multi-step task construal may require external structure beyond a single prompt.

\subsection{Limitations and Future Work}
\label{subsec:limitations_future}

The present study focuses on three English-language benchmark tasks with annotations suited to measuring frame activation, salience control, and construal selection. This design supports controlled comparison across prompt conditions, but it leaves open how the same framework extends to open-ended generation, dialogue, tool use, domain-specific applications, and multilingual settings. The experiments also use a limited set of manually designed prompts, so future work should test paraphrased prompt families and broader prompt distributions to separate semantic category effects from wording-specific effects.

The framework can also be extended beyond prompt-only interventions. Context-level experiments could vary retrieval quality, sentence ordering, distractor density, and evidence filtering to test how salience control depends on the information environment. Harness-level experiments could introduce decomposition, routing, verification, or tool use to test whether task construal becomes more stable when the surrounding system enforces intermediate structure. Combining these behavioral evaluations with representation-level analyses would further clarify how semantic control is implemented both inside language models and in the systems built around them.

\section{Conclusion}
\label{sec:conclusion}

We developed a cognitive-semantic account of prompting as semantic control. On this view, prompts are not only instructions for producing outputs, but linguistic conditions that shape how a fixed model interprets an input, foregrounds information, and organizes a task. The three experiments show that these effects are measurable across distinct forms of behavior: frame-oriented prompts shift semantic relation judgments in natural language inference, salience-oriented prompts change evidence use in claim verification, and construal-oriented prompts affect answer--support organization in multi-hop question answering. These differences suggest that prompt effects should be analyzed not only by performance gains, but by the direction and structure of the behavioral movement they induce. The resulting framework places prompt engineering within a broader stack of semantic control, where prompts guide interpretive frames, contexts shape evidential salience, and harnesses support task-level organization.

\section*{Broader Impact Statement}

This paper examines how natural-language prompts shape large language model behavior through frame activation, salience control, and construal selection. Its intended contribution is diagnostic. By treating prompts as semantic control conditions, the framework encourages researchers and practitioners to evaluate not only whether a prompt improves aggregate performance, but also how it moves model behavior across judgments, evidence use, and answer--support organization. This can make prompted systems easier to inspect in settings where the basis, framing, and structure of an output matter alongside correctness.

The same capacity creates risks. Prompts that shift labels, foreground evidence, or alter task construal can also be used to steer models toward biased judgments, selective rationales, misleading explanations, or unwarranted confidence. These risks are especially relevant in high-stakes domains such as education, hiring, finance, healthcare, and legal decision support, where changes in framing or evidence selection may affect how users interpret and trust model outputs. The framework should therefore be used to audit prompt effects, not to make model responses more persuasive without scrutiny.

The results also suggest that responsible prompt use should not rely on wording alone. Because semantic control differs across models, tasks, and levels of organization, prompts should be paired with careful context construction, structured interfaces, evidence checks, and task-specific validation when reliability matters. The broader impact of the paper is thus to provide concepts and measurements for identifying how prompts influence model behavior before prompted systems are used in consequential settings.

\section*{Reproducibility}

All experiments reported in this paper can be reproduced using the notebook \texttt{cognitive\_semantics.ipynb}, which is provided as supplementary material. The notebook contains the experimental prompts, dataset preprocessing steps, model-loading procedures, output parsing code, metric definitions, and table-generation scripts used for the reported results. We also include the random seeds and sampling procedures needed to reproduce the benchmark subsets and bootstrap confidence intervals.

\bibliography{main}
\bibliographystyle{tmlr}

\appendix

\section{Metric Definitions}
\label{app:metric_definitions}

This appendix provides the full metric definitions used in the experiments. The main text reports the prompt-effect diagnostics most directly tied to the cognitive-semantic analysis, while this appendix includes the standard task-performance, evidence-selection, and answer-support metrics used to compute the reported tables.

\subsection{Metrics for Experiment 1}

\begin{description}
    \item[Accuracy.]
    Overall correctness of the model's natural language inference predictions. A prediction is counted as correct when the predicted label matches the gold label among entailment, contradiction, and neutral.

    \item[Macro-F1.]
    Class-balanced classification performance across the three natural language inference labels. F1 is computed separately for entailment, contradiction, and neutral, and the three class-level scores are averaged.

    \item[Class-wise recall.]
    Recall computed separately for each target label. This metric is used to examine whether a prompt frame increases sensitivity to the semantic relation it foregrounds.

    \item[FrameSens.]
    The fraction of examples whose predicted label changes under at least one prompt frame. This metric captures whether changing the frame instruction induces any observable movement in the model's judgment for the same input.

    \item[FrameRecGain.]
    The average recall gain of each target frame relative to the neutral frame. For each target label, we compare recall under the corresponding frame prompt against recall under the neutral prompt, and then average the gains across target labels.

    \item[BiasShift.]
    The change in prediction rate for the frame-compatible label. This metric measures whether a prompt frame increases the model's tendency to choose the label emphasized by that frame, regardless of whether the prediction is correct.
\end{description}

\subsection{Metrics for Experiment 2}

\begin{description}
    \item[LabelAcc.]
    Correctness of the claim-verification label. A prediction is counted as correct when the predicted label matches the gold label among SUPPORTS, REFUTES, and NOT ENOUGH INFO.

    \item[Macro-F1.]
    Class-balanced classification performance across SUPPORTS, REFUTES, and NOT ENOUGH INFO. This metric complements label accuracy by reducing the influence of class imbalance.

    \item[EvidF1.]
    Example-level F1 overlap between the selected evidence sentence identifiers and the gold evidence sentence identifiers. This metric measures whether the model selects the evidence that supports the verification decision.

    \item[Joint.]
    Joint label-evidence success. An example is counted as successful when the predicted label is correct and the selected evidence satisfies the evidence-overlap criterion. This metric captures whether the model combines correct classification with adequate evidence grounding.

    \item[Leak.]
    The fraction of selected evidence sentences that are not gold evidence sentences. This metric measures distractor leakage, or the extent to which the model selects irrelevant candidate evidence.

    \item[SalGain.]
    The change in evidence F1 under the evidence-first prompt relative to the neutral verification prompt. This metric captures whether foregrounding evidence selection improves evidence grounding.

    \item[JointGain.]
    The change in joint label-evidence success under the evidence-first prompt relative to the neutral verification prompt. This metric measures whether evidence foregrounding improves the coupling between the final label and the selected evidence.

    \item[LeakReduction.]
    The reduction in evidence leakage under the evidence-first prompt relative to the neutral verification prompt. Positive values indicate that the evidence-first prompt reduces the selection of non-gold evidence sentences.

    \item[Recall gain.]
    Class-specific recall movement induced by support- or refutation-oriented prompts. Support recall gain compares SUPPORTS recall under the support-salience prompt against the neutral prompt, while refutation recall gain compares REFUTES recall under the refutation-salience prompt against the neutral prompt.

    \item[Bias shift.]
    Class-specific prediction-rate movement induced by support- or refutation-oriented prompts. Support bias shift measures the increase in SUPPORTS prediction rate under the support-salience prompt, while refutation bias shift measures the increase in REFUTES prediction rate under the refutation-salience prompt.
\end{description}

\subsection{Metrics for Experiment 3}

\begin{description}
    \item[AnsEM.]
    Exact match between the predicted answer and the gold answer after standard answer normalization. This metric measures whether the final answer string exactly matches the reference answer.

    \item[AnsF1.]
    Token-level F1 between the predicted answer and the gold answer after standard answer normalization. This metric gives partial credit when the predicted answer overlaps with the gold answer but is not an exact match.

    \item[SupF1.]
    Example-level F1 overlap between the selected supporting sentence identifiers and the gold supporting sentence identifiers. This metric measures whether the model identifies the facts needed to support the answer.

    \item[JointF1.]
    Product-based coupling of answer quality and supporting-fact quality. For each example, JointF1 is computed as the product of AnsF1 and SupF1, so high values require both a good answer and good supporting-fact selection.

    \item[AnsEvidCouple.]
    The rate at which sufficiently correct answers are paired with sufficiently correct supporting facts. In the experiments, this is computed by checking whether examples with high answer F1 also reach the supporting-fact F1 threshold.

    \item[TypeMatchedConGain.]
    The gain from using the prompt matched to the question type relative to the direct-answer baseline. Bridge prompts are matched to bridge questions, and comparison prompts are matched to comparison questions.

    \item[Answer/support gains.]
    Separate changes in answer quality, supporting-fact selection, and support-set size relative to the direct-answer prompt. These diagnostics distinguish whether a construal prompt changes the final answer, the selected support structure, or the amount of support selected.
\end{description}

\section{Prompt Templates and Output Schemas}
\label{app:prompt_templates}

\subsection{Experiment 1: Prompt Frames in Natural Language Inference}
\label{app:prompt_templates_exp1}

Experiment~1 used a natural language inference format built around a common premise--hypothesis judgment. The manipulation was introduced only through the frame instruction: the premise, hypothesis, allowed label set, output schema, and decoding procedure were held fixed across conditions. The model was required to return a single JSON object containing one label.

The fixed system instruction was as follows.

\begin{promptbox}[System prompt.]
\ttfamily
You are a careful natural language inference classifier.\\
Return only the requested JSON object. Do not explain your reasoning.
\end{promptbox}

The four prompt conditions differed only in the frame instruction inserted into the shared template.

\begin{promptbox}[Frame instructions.]
\ttfamily
Neutral frame:\\
Determine whether the hypothesis is entailed by, contradicts, or is neutral with respect to the premise.\\
\\
Entailment frame:\\
Focus on whether the premise provides enough support to make the hypothesis true.\\
\\
Contradiction frame:\\
Focus on whether the premise contains information that makes the hypothesis false.\\
\\
Uncertainty frame:\\
Focus on whether the premise leaves the hypothesis underdetermined.
\end{promptbox}

The shared user prompt template was as follows.

\begin{promptbox}[User prompt template.]
\ttfamily
Task: Natural Language Inference.\\
\\
Frame instruction:\\
\{frame instruction\}\\
\\
Premise:\\
\{premise\}\\
\\
Hypothesis:\\
\{hypothesis\}\\
\\
Allowed labels:\\
- entailment: the hypothesis follows from the premise.\\
- contradiction: the hypothesis conflicts with the premise.\\
- neutral: the premise does not determine whether the hypothesis is true or false.\\
\\
Return exactly one JSON object and no other text.\\
Schema:\\
\{"label": "entailment"\} OR \{"label": "contradiction"\} OR \{"label": "neutral"\}
\end{promptbox}

The expected output schema was therefore fixed across all four conditions.

\begin{promptbox}[Output schema.]
\ttfamily
\{"label": "entailment"\}\\
\\
or\\
\\
\{"label": "contradiction"\}\\
\\
or\\
\\
\{"label": "neutral"\}
\end{promptbox}

During parsing, a response was counted as a valid JSON response only when a JSON object with a usable \texttt{label} field was returned. Label extraction was also recorded separately, allowing non-JSON responses that nevertheless contained an identifiable label to be tracked as extraction successes but JSON-format failures. Invalid or unparseable labels were treated as invalid predictions for metric computation.

\subsection{Experiment 2: Evidence Foregrounding in Claim Verification}
\label{app:prompt_templates_exp2}

Experiment~2 used a claim-verification format built around a fixed claim and a fixed set of candidate evidence sentences. The manipulation was introduced only through the foregrounding instruction: the claim, candidate evidence set, allowed labels, output schema, and evidence-selection rules were held fixed across conditions. The model was required to return a single JSON object containing both a verification label and a list of selected evidence sentence identifiers.

The fixed system instruction was as follows.

\begin{promptbox}[System prompt.]
\ttfamily
You are a careful fact verification classifier.\\
Use only the provided evidence sentences.\\
Return only the requested JSON object. Do not explain your reasoning.
\end{promptbox}

The four prompt conditions differed only in the instruction inserted into the shared template.

\begin{promptbox}[Foregrounding instructions.]
\ttfamily
Neutral verification:\\
Verify the claim using the provided evidence sentences.\\
\\
Evidence-first salience:\\
First identify the minimal evidence sentences directly relevant to the claim.\\
Then decide whether the claim is SUPPORTS, REFUTES, or NOT ENOUGH INFO.\\
\\
Support salience:\\
Focus on evidence that could support the claim.\\
If the evidence directly supports the claim, choose SUPPORTS; otherwise choose the appropriate label.\\
\\
Refutation salience:\\
Focus on evidence that could refute the claim.\\
If the evidence directly contradicts the claim, choose REFUTES; otherwise choose the appropriate label.
\end{promptbox}

The candidate evidence sentences were rendered as indexed sentence identifiers. Each identifier could optionally include source metadata when available.

\begin{promptbox}[Candidate evidence format.]
\ttfamily
sent\_0 [\{title\}, line \{line\}]: \{evidence sentence\}\\
sent\_1 [\{title\}, line \{line\}]: \{evidence sentence\}\\
sent\_2 [\{title\}, line \{line\}]: \{evidence sentence\}\\
...\\
sent\_7 [\{title\}, line \{line\}]: \{evidence sentence\}
\end{promptbox}

The shared user prompt template was as follows.

\begin{promptbox}[User prompt template.]
\ttfamily
Task: FEVER-style claim verification.\\
\\
Instruction:\\
\{condition instruction\}\\
\\
Claim:\\
\{claim\}\\
\\
Candidate evidence sentences:\\
\{candidate evidence block\}\\
\\
Allowed labels:\\
- SUPPORTS: the provided evidence supports the claim.\\
- REFUTES: the provided evidence contradicts the claim.\\
- NOT ENOUGH INFO: the provided evidence is insufficient to verify or refute the claim.\\
\\
Return exactly one JSON object and no other text.\\
Schema:\\
\{"label": "SUPPORTS", "evidence\_ids": ["sent\_0"]\}\\
or\\
\{"label": "REFUTES", "evidence\_ids": ["sent\_1"]\}\\
or\\
\{"label": "NOT ENOUGH INFO", "evidence\_ids": []\}\\
\\
Rules:\\
- evidence\_ids must be selected only from the candidate sentence ids shown above.\\
- Use the minimal evidence set.\\
- For NOT ENOUGH INFO, use an empty evidence\_ids list unless a sentence is relevant but insufficient.
\end{promptbox}

The expected output schema was therefore fixed across all four conditions.

\begin{promptbox}[Output schema.]
\ttfamily
\{"label": "SUPPORTS", "evidence\_ids": ["sent\_0"]\}\\
\\
or\\
\\
\{"label": "REFUTES", "evidence\_ids": ["sent\_1"]\}\\
\\
or\\
\\
\{"label": "NOT ENOUGH INFO", "evidence\_ids": []\}
\end{promptbox}

During parsing, a response was counted as a valid JSON response when a JSON object with a usable label field was returned. Evidence identifiers were extracted from the returned evidence field when available and were filtered to the candidate sentence identifiers shown in the prompt. Label extraction and evidence extraction were recorded separately, allowing non-JSON responses with identifiable labels or sentence identifiers to be tracked as extraction successes but JSON-format failures. Invalid or unparseable labels were treated as invalid predictions for metric computation.

\subsection{Experiment 3: Task Construal in Multi-Hop Question Answering}
\label{app:prompt_templates_exp3}

Experiment~3 used a multi-hop question answering format built around a fixed question and a fixed set of candidate context sentences. The manipulation was introduced only through the construal instruction: the question, candidate context set, output schema, and supporting-sentence selection rules were held fixed across conditions. The model was required to return a single JSON object containing both a final answer and a list of selected supporting sentence identifiers.

The fixed system instruction was as follows.

\begin{promptbox}[System prompt.]
\ttfamily
You are a careful multi-hop question answering system.\\
Use only the provided context sentences.\\
Return only the requested JSON object. Do not explain your reasoning.
\end{promptbox}

The four prompt conditions differed only in the construal instruction inserted into the shared template.

\begin{promptbox}[Construal instructions.]
\ttfamily
Direct-answer construal:\\
Answer the question based on the provided context sentences.\\
\\
Supporting-facts-first construal:\\
First identify the supporting facts needed to answer the question.\\
Then give the final answer.\\
\\
Bridge-reasoning construal:\\
Interpret the question as requiring a bridge between multiple facts.\\
Identify the intermediate entity or fact before answering.\\
\\
Comparison construal:\\
Interpret the question as requiring comparison between relevant entities.\\
Identify the compared entities and the facts used for comparison before answering.
\end{promptbox}

The candidate context sentences were rendered as indexed sentence identifiers. Each identifier included the source title and the original sentence index when available.

\begin{promptbox}[Candidate context format.]
\ttfamily
sent\_0 [\{title\}, sentence \{sent\_id\}]: \{context sentence\}\\
sent\_1 [\{title\}, sentence \{sent\_id\}]: \{context sentence\}\\
sent\_2 [\{title\}, sentence \{sent\_id\}]: \{context sentence\}\\
...\\
sent\_13 [\{title\}, sentence \{sent\_id\}]: \{context sentence\}
\end{promptbox}

The shared user prompt template was as follows.

\begin{promptbox}[User prompt template.]
\ttfamily
Task: HotpotQA-style multi-hop question answering.\\
\\
Instruction:\\
\{condition instruction\}\\
\\
Question:\\
\{question\}\\
\\
Candidate context sentences:\\
\{candidate context block\}\\
\\
Return exactly one JSON object and no other text.\\
Schema:\\
\{"answer": "final answer string", "supporting\_sentence\_ids": ["sent\_0", "sent\_3"]\}\\
\\
Rules:\\
- Use only the candidate context sentences shown above.\\
- supporting\_sentence\_ids must be selected only from the candidate sentence ids shown above.\\
- Use the minimal supporting sentence set needed for the answer.\\
- If the answer is yes or no, return exactly "yes" or "no".
\end{promptbox}

The expected output schema was therefore fixed across all four conditions.

\begin{promptbox}[Output schema.]
\ttfamily
\{"answer": "final answer string", "supporting\_sentence\_ids": ["sent\_0", "sent\_3"]\}
\end{promptbox}

During parsing, a response was counted as a valid JSON response when a JSON object with a usable answer field was returned. Supporting sentence identifiers were extracted from the returned supporting-sentence field when available and were filtered to the candidate sentence identifiers shown in the prompt. Answer extraction and support extraction were recorded separately, allowing non-JSON responses with identifiable answers or sentence identifiers to be tracked as extraction successes but JSON-format failures. Empty or unparseable answers were treated as missing predictions for answer evaluation.

\section{Additional Experiment Confidence Intervals}
\label{app:experiment_confidence_intervals}

This appendix reports the 95\% bootstrap confidence intervals corresponding to the point estimates reported in Section~\ref{sec:experiments}. The main text reports point estimates to keep the experimental results readable, while the intervals below provide uncertainty estimates for the main aggregate metrics.

\begin{table*}[ht]
\centering
\small
\setlength{\tabcolsep}{4pt}
\caption{Bootstrap confidence intervals for Experiment 1.}
\label{tab:app_exp1_ci}

\begin{subtable}{\textwidth}
\centering
\caption{Prompt-condition confidence intervals.}
\begin{tabular}{llcc}
\toprule
Model & Condition & Accuracy & Macro-F1 \\
\midrule
\multirow{4}{*}{SmolLM2-1.7B}
& contradiction & [0.416, 0.500] & [0.332, 0.397] \\
& entailment    & [0.338, 0.422] & [0.226, 0.297] \\
& neutral       & [0.430, 0.520] & [0.390, 0.481] \\
& uncertainty   & [0.378, 0.464] & [0.328, 0.406] \\
\midrule
\multirow{4}{*}{Qwen3-4B}
& contradiction & [0.832, 0.892] & [0.832, 0.893] \\
& entailment    & [0.796, 0.864] & [0.798, 0.864] \\
& neutral       & [0.824, 0.888] & [0.827, 0.889] \\
& uncertainty   & [0.716, 0.792] & [0.719, 0.793] \\
\midrule
\multirow{4}{*}{Mistral-7B}
& contradiction & [0.670, 0.748] & [0.660, 0.742] \\
& entailment    & [0.572, 0.658] & [0.504, 0.568] \\
& neutral       & [0.584, 0.668] & [0.543, 0.623] \\
& uncertainty   & [0.478, 0.564] & [0.408, 0.469] \\
\bottomrule
\end{tabular}
\end{subtable}

\vspace{0.75em}

\begin{subtable}{\textwidth}
\centering
\caption{Frame-level confidence intervals.}
\begin{tabular}{lcc}
\toprule
Model & FrameSens & FrameRecGain \\
\midrule
SmolLM2-1.7B & [0.528, 0.614] & [0.143, 0.221] \\
Qwen3-4B    & [0.208, 0.280] & [0.060, 0.106] \\
Mistral-7B  & [0.306, 0.394] & [0.117, 0.180] \\
\bottomrule
\end{tabular}
\end{subtable}

\end{table*}

\begin{table*}[ht]
\centering
\small
\setlength{\tabcolsep}{3pt}
\caption{Bootstrap confidence intervals for Experiment 2 condition-level results.}
\label{tab:app_exp2_condition_ci}
\resizebox{\textwidth}{!}{
\begin{tabular}{llccccc}
\toprule
Model & Condition & LabelAcc & Macro-F1 & EvidF1 & Joint & Leak \\
\midrule
\multirow{4}{*}{SmolLM2-1.7B}
& neutral        & [0.398, 0.484] & [0.307, 0.367] & [0.082, 0.124] & [0.020, 0.052] & [0.886, 0.931] \\
& evidence-first & [0.486, 0.572] & [0.401, 0.456] & [0.111, 0.147] & [0.010, 0.036] & [0.898, 0.928] \\
& support        & [0.342, 0.430] & [0.236, 0.300] & [0.067, 0.115] & [0.022, 0.056] & [0.883, 0.935] \\
& refutation     & [0.476, 0.564] & [0.386, 0.444] & [0.103, 0.142] & [0.014, 0.040] & [0.887, 0.925] \\
\midrule
\multirow{4}{*}{Qwen3-4B}
& neutral        & [0.894, 0.942] & [0.895, 0.942] & [0.883, 0.929] & [0.884, 0.934] & [0.010, 0.036] \\
& evidence-first & [0.876, 0.928] & [0.876, 0.927] & [0.859, 0.908] & [0.868, 0.922] & [0.002, 0.018] \\
& support        & [0.870, 0.922] & [0.867, 0.921] & [0.864, 0.911] & [0.860, 0.914] & [0.006, 0.028] \\
& refutation     & [0.884, 0.934] & [0.885, 0.934] & [0.892, 0.933] & [0.874, 0.924] & [0.018, 0.048] \\
\midrule
\multirow{4}{*}{Mistral-7B}
& neutral        & [0.558, 0.638] & [0.500, 0.579] & [0.562, 0.641] & [0.552, 0.634] & [0.002, 0.020] \\
& evidence-first & [0.566, 0.652] & [0.509, 0.581] & [0.578, 0.662] & [0.564, 0.650] & [0.002, 0.016] \\
& support        & [0.570, 0.652] & [0.502, 0.568] & [0.574, 0.658] & [0.566, 0.648] & [0.004, 0.024] \\
& refutation     & [0.498, 0.588] & [0.469, 0.563] & [0.475, 0.565] & [0.456, 0.548] & [0.062, 0.110] \\
\bottomrule
\end{tabular}
}
\end{table*}

\begin{table*}[ht]
\centering
\small
\setlength{\tabcolsep}{3pt}
\caption{Bootstrap confidence intervals for Experiment 2 salience-level results.}
\label{tab:app_exp2_salience_ci}

\begin{subtable}{\textwidth}
\centering
\caption{Salience-level aggregate metrics.}
\begin{tabular}{lccc}
\toprule
Model & SalGain & JointGain & LeakRed. \\
\midrule
SmolLM2-1.7B
& [\hphantom{-}0.007, \hphantom{-}0.046]
& [-0.030, 0.002]
& [-0.024, 0.015] \\
Qwen3-4B
& [-0.041, -0.005]
& [-0.036, 0.004]
& [\hphantom{-}0.004, 0.022] \\
Mistral-7B
& [\hphantom{-}0.001, \hphantom{-}0.040]
& [-0.004, 0.038]
& [-0.006, 0.010] \\
\bottomrule
\end{tabular}
\end{subtable}

\vspace{0.75em}

\begin{subtable}{\textwidth}
\centering
\caption{Recall, bias, and verification diagnostics.}
\resizebox{\textwidth}{!}{
\begin{tabular}{lccccccc}
\toprule
\multicolumn{1}{c}{\multirow{2}{*}{Model}} &
\multicolumn{2}{c}{Recall Gain} &
\multicolumn{2}{c}{Bias Shift} &
\multicolumn{3}{c}{Verification} \\
\cmidrule(lr){2-3}
\cmidrule(lr){4-5}
\cmidrule(lr){6-8}
& 
\multicolumn{1}{c}{SUPPORTS} &
\multicolumn{1}{c}{REFUTES} &
\multicolumn{1}{c}{SUPPORTS} &
\multicolumn{1}{c}{REFUTES} &
\multicolumn{1}{c}{SalGain} &
\multicolumn{1}{c}{JointGain} &
\multicolumn{1}{c}{LeakRed.} \\
\midrule
SmolLM2-1.7B
& [\hphantom{-}0.006, 0.070]
& [0.000, 0.000]
& [0.112, 0.176]
& [0.000, 0.000]
& [\hphantom{-}0.011, \hphantom{-}0.068]
& [-0.044, \hphantom{-}0.003]
& [-0.036, 0.024] \\
Qwen3-4B
& [-0.006, 0.045]
& [0.006, 0.050]
& [0.008, 0.034]
& [0.018, 0.052]
& [-0.076, -0.026]
& [-0.075, -0.023]
& [\hphantom{-}0.000, 0.009] \\
Mistral-7B
& [\hphantom{-}0.054, 0.164]
& [0.133, 0.253]
& [0.020, 0.060]
& [0.104, 0.164]
& [\hphantom{-}0.002, \hphantom{-}0.058]
& [-0.006, \hphantom{-}0.056]
& [-0.006, 0.014] \\
\bottomrule
\end{tabular}
}
\end{subtable}

\end{table*}

\begin{table*}[ht]
\centering
\small
\setlength{\tabcolsep}{3pt}
\caption{Bootstrap confidence intervals for Experiment 3 condition-level results.}
\label{tab:app_exp3_condition_ci}
\resizebox{\textwidth}{!}{
\begin{tabular}{llcccccc}
\toprule
Model & Condition & AnsEM & AnsF1 & SupF1 & JointF1 & Coupling & PredSup \\
\midrule
\multirow{4}{*}{SmolLM2-1.7B}
& direct        & [0.020, 0.080] & [0.020, 0.080] & [0.159, 0.225] & [0.003, 0.023] & [0.000, 0.616] & [2.730, 3.765] \\
& support-first & [0.025, 0.080] & [0.025, 0.080] & [0.151, 0.220] & [0.003, 0.020] & [0.000, 0.500] & [2.725, 3.655] \\
& bridge        & [0.020, 0.085] & [0.020, 0.085] & [0.162, 0.230] & [0.001, 0.020] & [0.000, 0.500] & [2.779, 3.790] \\
& comparison    & [0.025, 0.080] & [0.025, 0.080] & [0.147, 0.213] & [0.001, 0.017] & [0.000, 0.500] & [2.560, 3.455] \\
\midrule
\multirow{4}{*}{Qwen3-4B}
& direct        & [0.530, 0.665] & [0.668, 0.775] & [0.740, 0.807] & [0.524, 0.626] & [0.881, 0.970] & [1.710, 1.865] \\
& support-first & [0.530, 0.660] & [0.662, 0.768] & [0.758, 0.824] & [0.524, 0.624] & [0.890, 0.975] & [1.740, 1.870] \\
& bridge        & [0.515, 0.655] & [0.667, 0.777] & [0.750, 0.822] & [0.529, 0.630] & [0.882, 0.969] & [1.775, 1.900] \\
& comparison    & [0.460, 0.590] & [0.612, 0.721] & [0.738, 0.812] & [0.485, 0.586] & [0.867, 0.964] & [1.705, 1.835] \\
\midrule
\multirow{4}{*}{Mistral-7B}
& direct        & [0.360, 0.495] & [0.476, 0.601] & [0.599, 0.677] & [0.332, 0.431] & [0.867, 0.970] & [1.910, 2.015] \\
& support-first & [0.330, 0.470] & [0.466, 0.587] & [0.586, 0.664] & [0.314, 0.411] & [0.852, 0.966] & [1.925, 2.025] \\
& bridge        & [0.365, 0.500] & [0.477, 0.603] & [0.581, 0.660] & [0.324, 0.429] & [0.857, 0.971] & [1.915, 2.010] \\
& comparison    & [0.350, 0.485] & [0.477, 0.601] & [0.592, 0.674] & [0.313, 0.412] & [0.776, 0.920] & [1.905, 1.995] \\
\bottomrule
\end{tabular}
}
\end{table*}

\begin{table*}[ht]
\centering
\small
\setlength{\tabcolsep}{3pt}
\caption{Bootstrap confidence intervals for Experiment 3 construal-level results.}
\label{tab:app_exp3_construal_ci}
\resizebox{\textwidth}{!}{
\begin{tabular}{lcccccccc}
\toprule
Model & SupFirst & TypeMatch & AnsGain & SupGain & SuppExp. & BridgeOnly & CompOnly & CoupleGain \\
\midrule
SmolLM2-1.7B
& [-0.005, 0.000]
& [-0.009, 0.000]
& [\hphantom{-}0.000, 0.000]
& [-0.020, 0.005]
& [-0.680, 0.150]
& [\hphantom{-}0.000, 0.000]
& [-0.017, 0.000]
& [-0.333, \hphantom{-}0.000] \\
Qwen3-4B
& [-0.033, 0.031]
& [-0.050, 0.020]
& [-0.065, 0.005]
& [-0.015, 0.031]
& [-0.010, 0.105]
& [-0.022, 0.060]
& [-0.100, 0.004]
& [-0.047, \hphantom{-}0.031] \\
Mistral-7B
& [-0.035, 0.001]
& [-0.033, 0.029]
& [-0.004, 0.056]
& [-0.052, 0.018]
& [-0.055, 0.045]
& [-0.037, 0.028]
& [-0.050, 0.051]
& [-0.129, -0.007] \\
\bottomrule
\end{tabular}
}
\end{table*}

\end{document}